\documentclass[letterpaper, 10 pt, conference]{ieeeconf}

\IEEEoverridecommandlockouts  

\usepackage{fix-cm}
\usepackage{etex}

\usepackage{dblfloatfix}

\usepackage{nag}

\makeatletter
\@ifpackageloaded{xcolor}{}{%
\usepackage[table,x11names,dvipsnames,svgnames]{xcolor}%
}
\makeatother

\usepackage{colortbl}

\usepackage{graphicx}
\usepackage{wrapfig}

\definecolorset{RGB}{lyft}{}{Red,194,39,36;Sunset,202,53,33;Orange,205,68,20;Amber,200,117,42;Yellow,242,169,52;Citron,186,188,44;Lime,112,159,33;Green,56,139,31;Mint,45,118,56;Teal,52,133,135;Cyan,60,132,202;Blue,55,94,248;Indigo,64,13,247;Purple,115,42,248;Pink,176,25,145;Rose,176,32,75}

\definecolorset{HTML}{h}{}{grapefruit,ED5565;bittersweet,FC6E51;sunflower,FFCE54;grass,A0D468;mint,48CFAD;aqua,4FC1E9;bluejeans,5D9CEC;lavender,AC92EC;pinkrose,EC87C0;lightgray,F5F7FA;mediumgray,CCD1D9;darkgray,656D78}

\usepackage{cite}

\usepackage{microtype}

\usepackage[american]{babel}

\usepackage{array}
\usepackage{multirow}
\usepackage{booktabs}
\usepackage{makecell} 

\newcommand{\myparagraph}[1]{\textbf{\emph{#1}}.}

\ifcsname labelindent\endcsname
\let\labelindent\relax
\fi
\usepackage[inline]{enumitem}

\usepackage{subfig}

\newenvironment{lenumerate}[2][]
{\begin{enumerate}[label=(#2\arabic*),leftmargin=0in,itemindent=0.35in,#1]}
{\end{enumerate}}

\setlist*[enumerate,1]{label={\itshape\arabic*)}}

\makeatletter
\newcommand{\paragraphswithstop}{%
\let\copyparagraph\paragraph%
\renewcommand\paragraph[1]{\copyparagraph{##1.}}%
}
\makeatother

\usepackage[framemethod=tikz]{mdframed}

\makeatletter
\def\namedlabel#1#2{\begingroup
  #2%
  \def\@currentlabel{#2}%
  \phantomsection\label{#1}\endgroup
}
\makeatother

\makeatletter
\def\namedlabelphantom#1#2{\begingroup
  \def\@currentlabel{#2}%
  \phantomsection\label{#1}\endgroup
}
\makeatother

\newcommand{\parunskip}{\bgroup\unskip\parfillskip=0pt \par\egroup}

\input{preamble/math}

\DeclarePairedDelimiter{\abs}{\lvert}{\rvert}

\DeclarePairedDelimiter{\norm}{\lVert}{\rVert}

\providecommand{\cG}{\mathcal{G}}

\providecommand{\cM}{\mathcal{M}}

\providecommand{\cV}{\mathcal{V}}

\usepackage{units}

  \newcommand{\newcolorlabel}[2]{%
  \expandafter\newcommand\csname #1\endcsname[1]{%
    \tikz[baseline]{\node[text=white,fill=#2,anchor=base,text height=1.3ex,text depth=0.1ex,font=\sffamily\bfseries]{##1}}}%
}

\newcommand{\newcommenter}[2]{%
  \expandafter\newcommand\csname #1\endcsname[1]{%
    \fcolorbox{#2}{#2}{\color{white}\textsf{\textbf{#1}}}
    {\color{#2}##1}}%
  \expandafter\newcommand\csname at#1\endcsname{%
    \fcolorbox{#2}{#2}{\color{white}\textsf{\textbf{@#1}}}
    {\color{#2}}}%
  \expandafter\newcommand\csname #1cite\endcsname[1]{%
    \csname #1\endcsname {[##1]}
  }%
  \expandafter\newcommand\csname #1ref\endcsname[1]{%
    \csname #1\endcsname {$\blacktriangleright$##1}
  }%
  \expandafter\newcommand\csname #1hl\endcsname[2]{%
    \colorbox{#2}{\color{white}\textsf{\textbf{#1}}}\sethlcolor{Azure2}\hl{##2}~%
    \expandafter\ifx\csname commentarrow\endcsname\relax$\leftarrow$\else \commentarrow[#2]\fi~%
    {\color{#2}##1}}%
  \expandafter\newcommand\csname #1st\endcsname[2]{%
    \colorbox{#2}{\color{white}\textsf{\textbf{#1}}}\sout{##2}~%
    \expandafter\ifx\csname commentarrow\endcsname\relax$\leftarrow$\else \commentarrow[#2]\fi~%
    {\color{#2}##1}}%
}
\newcommenter{TODO}{DodgerBlue1}
\newcommenter{rtron}{Green3}

\usepackage{comment}

\usepackage{pdfcomment}

\usepackage{soul}

\usepackage[normalem]{ulem}

\usepackage{csquotes}

\usepackage{suffix}

\usepackage{environ}

\makeatletter
\newsavebox{\boxifnotempty}
\newcommand{\displayifnotempty}[3]{\sbox\boxifnotempty{#2}\setbox0=\hbox{\usebox{\boxifnotempty}\unskip}%
  \ifdim\wd0=0pt
  \else
  #1\usebox{\boxifnotempty}#3%
  \fi%
}

\newcommand{\ifempty}[2]{\setbox0=\hbox{#1\unskip}%
  \ifdim\wd0=0pt%
  #2%
  \fi%
}

\newcommand{\ifnotempty}[2]{\setbox0=\hbox{#1\unskip}%
  \ifdim\wd0>0pt%
  #2%
  \fi%
}
\makeatother

\newcommand{\switchifempty}[3]{\sbox\boxifnotempty{#1}\setbox0=\hbox{\usebox{\boxifnotempty}\unskip}%
  \ifdim\wd0=0pt{}%
  #2%
  \else{}%
  #3%
  \usebox{\boxifnotempty}%
  \fi%
}

\makeatletter
\@ifundefined{chapter}{\usepackage{algorithm}}{\usepackage[chapter]{algorithm}}
\makeatother
\usepackage{algorithmicx}
\usepackage{algpseudocode}
\makeatother%

\usepackage{scrlfile}

\makeatletter
\newcommand*\newstoreddef[1]{
  \BeforeClosingMainAux{%
    \immediate\write\@auxout{%
      \string\restoredef{#1}{\csname #1\endcsname}%
    }%
  }%
}
\newcommand*{\restoredef}[2]{
  \expandafter\gdef\csname stored@#1\endcsname{#2}%
}
\newcommand*{\storeddef}[1]{
  \@ifundefined{stored@#1}{0}{\csname stored@#1\endcsname}%
}
\makeatother

\usepackage{pageslts}
\NewEnviron{tee}{\BODY\typeout{Marker Tee [start] ^^J \BODY ^^JMaker Tee [end]}}

\usepackage{cleveref}
\usepackage{nameref}

\usepackage{tikz}
\usetikzlibrary{calc}
\usetikzlibrary{matrix}
\usetikzlibrary{chains,scopes}
\usetikzlibrary{shapes.geometric}
\usetikzlibrary{arrows.meta}
\usetikzlibrary{decorations.markings}
\usetikzlibrary{decorations.pathreplacing}
\usetikzlibrary{backgrounds}

\tikzset{
  dim above/.style={to path={\pgfextra{
        \pgfinterruptpath
        \draw[>=latex,|->|] let
        \p1=($(\tikztostart)!1.5em!90:(\tikztotarget)$),
        \p2=($(\tikztotarget)!1.5em!-90:(\tikztostart)$)
        in(\p1) -- (\p2) node[pos=.5,sloped,above]{#1};
        \endpgfinterruptpath
      }
    }
  },
  dim double above/.style={to path={\pgfextra{
        \pgfinterruptpath
        \draw[>=latex,|->|] let
        \p1=($(\tikztostart)!3em!90:(\tikztotarget)$),
        \p2=($(\tikztotarget)!3em!-90:(\tikztostart)$)
        in(\p1) -- (\p2) node[pos=.5,sloped,above]{#1};
        \endpgfinterruptpath
      }
    }
  },
  dim below/.style={to path={\pgfextra{
        \pgfinterruptpath
        \draw[>=latex,|->|] let
        \p1=($(\tikztostart)!-1em!-90:(\tikztotarget)$),
        \p2=($(\tikztotarget)!-1em!90:(\tikztostart)$)
        in (\p1) -- (\p2) node[pos=.5,sloped,below]{#1};
        \endpgfinterruptpath
      }
    }
  },
}

\tikzset{
    right angle quadrant/.code={
        \pgfmathsetmacro\quadranta{{1,1,-1,-1}[#1-1]}     
        \pgfmathsetmacro\quadrantb{{1,-1,-1,1}[#1-1]}},
    right angle quadrant=1, 
    right angle length/.code={\def\rightanglelength{#1}},   
    right angle length=2ex, 
    right angle symbol/.style n args={3}{
        insert path={
            let \p0 = ($(#1)!(#3)!(#2)$) in     
                let \p1 = ($(\p0)!\quadranta*\rightanglelength!(#3)$), 
                \p2 = ($(\p0)!\quadrantb*\rightanglelength!(#2)$) in 
                let \p3 = ($(\p1)+(\p2)-(\p0)$) in  
            (\p1) -- (\p3) -- (\p2)
        }
    }
}

\newcommand{\pgfextractangle}[3]{%
    \pgfmathanglebetweenpoints{\pgfpointanchor{#2}{center}}
                              {\pgfpointanchor{#3}{center}}
    \global\let#1\pgfmathresult
}

\usetikzlibrary{shapes.arrows}
\newcommand{\commentarrow}[1][Azure4]{\tikz[baseline=-3pt]{\node[shape border uses incircle, fill=#1,rotate=180,single arrow, inner sep=1pt, minimum size=6pt, single arrow head extend=2pt]{};}}

\tikzset{ax/.style={-latex,line width=2pt}}

\tikzset{camera/.style={fill=Sienna1,fill opacity=0.5},%
image plane/.style={draw=RoyalBlue3,line width=2pt}}

\newcommand{\astar}{$\mathtt{A^*}$}

\newcommenter{todo}{Firebrick1}
\newcommenter{zili}{Firebrick1}
\newcommenter{sba}{Purple}

\title{\LARGE \bf
BoxMap: Efficient Structural Mapping and Navigation
}

\author{Zili Wang$^{}$, Christopher Allum$^{}$, Sean B. Andersson$^{}$, and Roberto Tron$^{}$
\thanks{Z.~Wang is with the Division of Systems Engineering, C.~Allum with the Department of Mechanical Engineering, and S.B.~Andersson and R.~Tron with the Division of Systems Engineering and the Department of Mechanical Engineering, Boston University, Boston, MA 02215, USA.
        {\tt\small \{zw2445,ctallum,sanderss,tron\}@bu.edu}.}%
\thanks{This work was supported in part by NSF FRR-2212051 and the Center for Information and Systems Engineering at Boston University.}
}

\begin{document}

\maketitle
\thispagestyle{empty}
\pagestyle{empty}

\begin{abstract}
    While humans can successfully navigate using abstractions, ignoring details that are irrelevant to the task at hand, most existing robotic applications require the maintenance of a detailed environment representation which consumes a significant amount of sensing, computing, and storage. These issues are particularly important in a resource-constrained setting with limited power budget.
  Deep learning methods can learn from prior experience to abstract knowledge of unknown environments, and use it to execute tasks (e.g., frontier exploration, object search,  or scene understanding) more efficiently.
  We propose BoxMap, a Detection-Transformer-based architecture that takes advantage of the structure of the sensed partial environment to update a topological graph of the environment as a set of semantic entities (e.g. rooms and doors) and their relations (e.g. connectivity). These predictions from low-level measurements can then be leveraged to achieve high-level goals with lower computational costs than methods based on detailed representations.
  As an example application, we consider a robot equipped with a 2-D laser scanner tasked with exploring a residential building.
  Our BoxMap representation scales quadratically with the number of rooms (with a small constant), resulting in significant savings over a full geometric map. Moreover, our high-level topological representation results in $30.9\%$ shorter trajectories in the exploration task with respect to a standard method. 
\end{abstract}

\section{INTRODUCTION}
Navigating complex and unknown environments to achieve various tasks is a remarkable ability shared by both humans and animals. This capability is derived in part from the ability to use past experiences to predict salient environmental features and identify the most effective strategies for success~\cite{GoalDrivenNavigation2018_EmbodiedNavigation, ObjectNav2018_graph, ObjectNav2020_winner}. Developing a similar ability for robots would be a valuable skill in real-world settings. While current approaches such as hierarchical abstractions utilizing semantic entities and physical relations have demonstrated success in long-term mapping, localization~\cite{rosinol2021kimera, bavle2023s, hughes2024foundations}, and planning~\cite{ravichandran2022hierarchical, kremer2023s}, they typically require extensive data collection via high-performance, resource-intensive sensors, alongside computationally demanding online algorithms. These requirements make such methods impractical for scenarios with limited resources, such as when using small-scale robots like the Harvard Ambulatory MicroRobot (HAMR)~\cite{Baisch2011} or during long-duration missions~\cite{Egerstedt2018}.

Motivated by these issues, in this paper we explore an end-to-end method we call BoxMap that transfers low-level raw measurements to topological maps (i.e., high-level semantic representations) of a structured environment by leveraging the predictability of its layout, such as indoor spaces comprised of polygonal rooms and with connecting doors.

Traditional approaches to generate a topological map of the environment are typically based either on pixel-wise semantic segmentation~\cite{bormann2016room, ijgi6070206,semanticMap2017} with the identification of physical relations between semantic entities~\cite{TopometricSemanticMaps2019}, or rely on the identification and clustering of wall entities~\cite{luperto2022robust, bavle2023s}. These methods do not take advantage of prior information about the environmental structure and the accuracy of the map relies on the post-processing procedures. Our previous work~\cite{wang2023more} proposed a multi-task learning architecture for multi-level abstractions; however, the topology generation was still based on ad-hoc post-processing of a low-level representation (occupancy maps).


With the development of deep learning, the prediction of polygonal instances typically involves two sequential steps
\begin{enumerate*}
  \item predicting bounding boxes containing each polygon based on anchors or proposals, and
  \item detecting polygon corners inside a representative bounding box, by using either a Convolutional Neural Network (CNN)-Recurrent Neural Network (RNN)~\cite{li2019topological} or CNN-Graph Convolutional Network (GCN) paradigm~\cite{ling2019fast, wei2021graph}.
\end{enumerate*}
The final results rely on many hand-designed components like a Non-Maximum Suppression (NMS) procedure or anchor generation.

To alleviate the effect of reliance on a hand-crafted process, an end-to-end method is desirable. A DEtection TRansformer (DETR) can directly predict the set of bounding boxes, with a CNN backbone (to extract image features), a transformer encoder-decoder architecture (to eliminate anchor generation) and a set-based global loss that forces unique predictions via bipartite matching (to eliminate NMS)~\cite{carion2020end}. Later work built upon a DETR to extract non-rectangular polygons by adding a polygon regression head and a corner classification head~\cite{hu2022polybuilding}. This approach, however, required multi-resolution representations and extra post-processing. Unlike these prior works that use bounding boxes to \textit{detect} objects, the high-level representation we develop here uses boxes as \textit{primitives} (i.e. building blocks) to represent an environment.

\begin{figure*}[t!]
  \centering
  \includegraphics[width=0.9\linewidth]{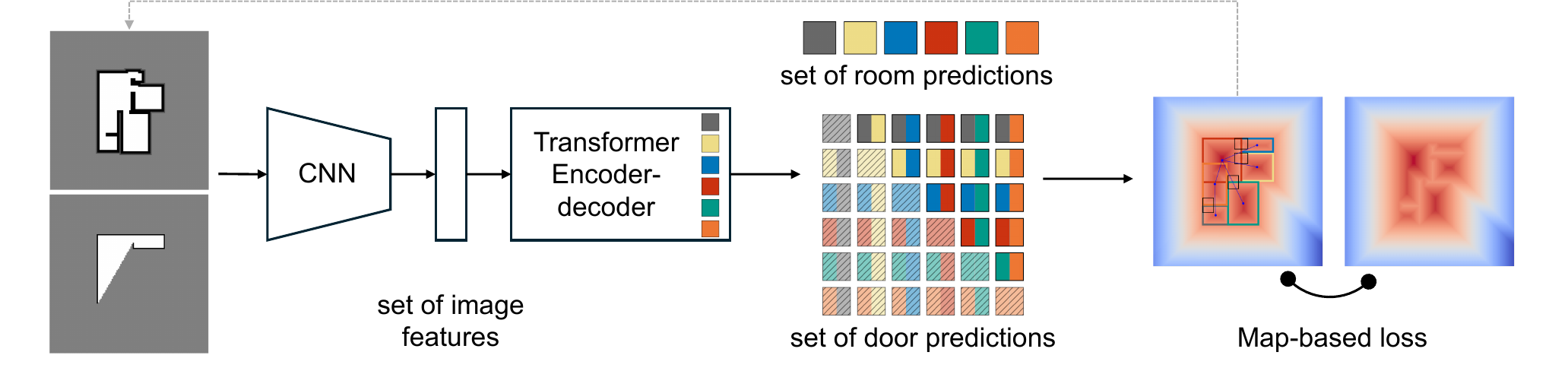}
  \caption{Architecture diagram: the TSDF from previous prediction is converted to an occupancy grid map, which is concatenated with the new laser measurement to be fed into a DETR-based model (a combination of CNN and transformer) to give multiple embeddings. Each embedding is used to predict a room box and each pair of embeddings is used to predict a door box. These boxes are transformed into a single TSDF, which is then compared with the ground truth to update the model.}
  \label{fig:dl_architecture}
\end{figure*}

As a concrete example for the approach we develop, we focus on an exploration task in an unknown environment. There are multiple approaches in the literature to achieve this particular task. Traditional methods typically build a geometric map from low-level representations (points, lines) using Simultaneous Localization and Mapping (SLAM), combined with a heuristic exploration method like frontier-based exploration~\cite{yamauchi1997frontier}. In more recent work, environmental priors are used to predict unknown regions based on assumed structure such as lines~\cite{MapPredictionExploration2021} or through the use of deep learning models~\cite{dl_frontier2019}. These predictions are then used to enhance exploration efficiency.
Alternative methods to improving exploration efficiency include predicting a directional hint to select a frontier point~\cite{pmlr-v87-stein18a, rl_frontier2020,arnob2023improving}.

\myparagraph{Paper contributions}
BoxMap models environments (rooms and doors) as boxes and their relations, and uses a DETR-like framework (Fig.~\ref{fig:dl_architecture}) to achieve end-to-end topological graph prediction from low-level measurements. Specifically:
\begin{itemize}[leftmargin=*]
  \item We create interpretable embeddings (box representations) for the prediction of rooms and their physical relations, via a CNN feature extractor followed by an encoder-decoder transformer that models room arrangements.
  \item We use losses based on the Truncated Signed Distance Function (TSDF) of predicted boxes to learn against the ground truth TSDF. Compared to losses based on vertices, our method is easier to train and avoids the necessity of manual labelling.
  \item We introduce a hierarchical loss that improves the detection of small but topologically important details (such as doors); the loss subtracts large objects (rooms) from a TSDF to highlight and focus on small ones.
\end{itemize}
The exploration example shows that our high-level graph representation enables
\begin{itemize}[leftmargin=*]
  \item semantic-level mapping by updating the graph with low-level measurements, yielding a significantly more compact memory footprint than a detailed map representation,
  \item higher-level reasoning of the environment that leads to structured inferences even beyond observed regions. The resulting graph-based decision making leads to shorter paths than standard frontier exploration methods.
\end{itemize}



\section{System Overview} \label{sec:objective}
We consider the problem of using a robot to explore an unknown environment from a random initial position with minimal map storage and total path traveled. We assume that:  \begin{enumerate*}
\item the robot is equipped with a laser range scanner;
\item the robot can detect the map alignment (i.e., the predominant direction of walls using, e.g.,  the methods in~\cite{Matteo2021structure,illingworth1988survey});
\item the robot has relatively accurate odometry.
\end{enumerate*}

The core idea of BoxMap is to leverage prior experience (a training set), and represent environments as graphs of boxes. The rest of this paper presents how our BoxMap representation is generated, and how it can be used for exploration.

\myparagraph{Graph Prediction and Mapping}
 BoxMap represents rectangular rooms as boxes with four corners,  non-rectangular rooms as  overlapping boxes (which we refer to as multi-boxes), and doorways as  square boxes connecting rooms. These are organized in a graph that is incrementally built from individual laser measurements via a learned model.

 At the robot position $p_t$, a laser measurement is acquired and converted to a local occupancy map, $M_t^{laser}$. This map is then concatenated with an occupancy grid map, $\hat{M}_{t-1}^{{topo}}$, instantiated from last step's topological graph, $\hat{\mathcal{G}}_{t-1}^{topo}$, to predict a new graph $\hat{\mathcal{G}}_t^{topo}$. 
 This step requires us to combine topological and occupancy map information; we use TSDF as a tool to bridge the gap between these two representations. Overall, the mapping update process is represented as
    \begin{equation}\label{eq:f_mapping_topo}
        \hat{\mathcal{G}}_t^{topo} = (\mathcal{V}, \mathcal{E}) = f_{map}(p_{t-1},p_t,{\hat{\mathcal{G}}_{t-1}^{topo}},M_t^{laser}),
    \end{equation}
    where $\mathcal{V}$ is the set of nodes, $\mathcal{E}$ is the set of edges showing the spatial connections between nodes, and $p_t$ is the position of the robot at time step $t$.

  \myparagraph{Graph-based exploration}
We use the graph $ \hat{\mathcal{G}}_t^{topo}$ to construct a planning graph $\hat{\mathcal{G}}_t^{nav}$ and then determine a frontier point $g_t$ to steer to. Here, we define a candidate frontier on $ \hat{\mathcal{G}}_t^{topo}$ as a room that is not visited. We then use the \astar{} algorithm to find a feasible trajectory from $p_t$ to $g_t$ on an occupancy grid map $M_t^{occ}$, which is accumulated on measurements up to the current time. Note that in our work, we keep our focus on the graph prediction, mapping and frontier selection, therefore maintaining $M_t^{occ}$ only in the \astar{} planning for simplicity. However, it is not necessary and in practice a local approach that uses only recent information should be sufficient. The graph is updated with a new measurement once the selected frontier has been reached, and the steps are iterated until a termination criteria is met.

\section{Graph Prediction and Mapping}
\subsection{BoxMap Network Architecture}
We use a DETR-based architecture as illustrated in Fig.~\ref{fig:dl_architecture} to implicitly represent prior structured information in the environment. It contains a CNN backbone, an encoder-decoder transformer, and task-specific prediction heads.

\subsubsection{Backbone}
A CNN backbone is first used to extract lower-resolution image feature maps. It takes the concatenated local maps $\hat{M}_{t}^{{topo}}$ and $M_{t}^{laser}$ and learns a common latent space representation. Each layer includes a downsampling convolution, followed by a standard convolution, both with Batch Normalization and Leaky ReLU. We use four layers with channel sizes $32$, $64$, $128$, and $256$.
\subsubsection{Transformer}
We use a standard transformer architecture to transform the feature maps into a set of embeddings. The transformer encoder processes the flattened image features and learned positional encodings as input and consists of 4 encoder layers. The transformer decoder is similarly composed of 4 decoder layers. Each encoder and decoder layer includes a multi-head attention mechanism and a feed-forward layer. The first decoder layer applies cross-attention, while the remaining layers use self-attention. The decoder produces $M$ embeddings for $M$ boxes in parallel, with box queries randomly initialized at the decoder’s input.

\subsubsection{Prediction heads}
\begin{figure}[tb]
  \centering
  \includegraphics[width=0.32\linewidth]{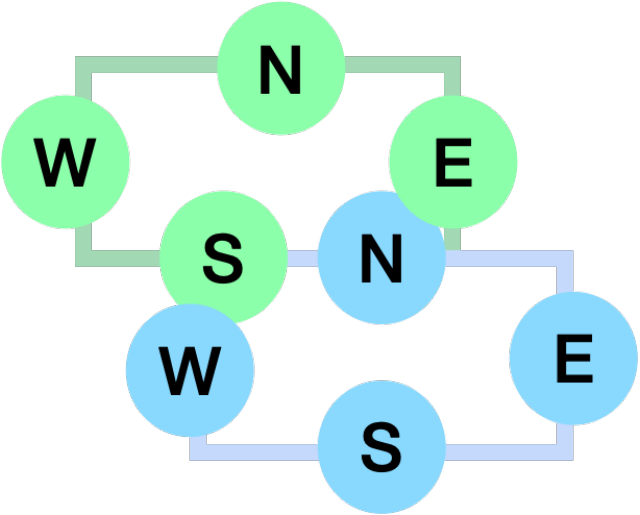}
\caption{The connectivity between two room boxes is validated by comparing the edges N, S, W, E of one room (in green) to the edges S, N, E, W of the other room (in blue). An entry will appear in the room box adjacency matrix if corresponding edges overlap, in this case green S and blue N.
}
\label{fig:room_adj}
\end{figure}

We define a box by its top left and bottom right coordinates $b^i=((x_0,y_0),(x_1,y_1))$, together with the class $q^i$. Embeddings from the transformer output are adopted to generate the prediction heads for rooms and doors.
Each room head predicts:
\begin{enumerate*}
    \item A class specifying the box existence.
    \item The room box coordinates.
\end{enumerate*}
%
From the predicted room box coordinates, we create an adjacency matrix between boxes (see Fig.~\ref{fig:room_adj}).
Two additional door prediction heads generate door predictions from all possible outer sums of pairs of embeddings. Each door head predicts:
\begin{enumerate*}
  \item A class specifying the door existence. This prediction is masked by the aforementioned adjacency matrix.
  \item The door box coordinates relative to the corresponding edge. 
\end{enumerate*}

\subsection{Parametric TSDFs for Structured Shapes}\label{subsec:sdf_from_box}
Although there are multiple ways to represent a non-rectangular room as a multi-box, the TSDF of the shape is unique. 
Therefore, we propose to define an analytic map (based on Rectified Linear Units, $ReLU(x) = \max(0,x)$) from box coordinates $b^i$ to a TSDF, and then base the training loss on TSDFs (\Cref{sec:loss function}).
In the following, $p = (x,y)$ represents a point in the environment, and $\gamma$ defines the truncation value for a TSDF.

\myparagraph{Box Representation}
The TSDF representation of a $2D$ rectangle in the $L_\infty$ norm is represented as the distance to the nearest wall along either the $x$ or $y$ axis,
\begin{equation}\label{eq:tsdf_rectApprox}
    f_{2D}(p) = \min(f_{1D,x},f_{1D,y});
\end{equation}
the distance along one axis can be represented by multiple $ReLU$ functions:
\begin{multline}
    f_{1D,x}(x_0,x_1) = \min(ReLU(x-x_0+\gamma)
    -ReLU(x-x_0-\gamma)\\-\gamma, -ReLU(x-x_1+\gamma)
    +ReLU(x-x_1-\gamma)+\gamma);\label{eq:sdf_1d}
\end{multline}
$f_{1D,y}$ is obtained by substituting $y$ for $x$ in~\eqref{eq:sdf_1d}.

\begin{figure}[tb]
  \centering
  \vspace{-2mm}
  \subfloat[Room merge]{
  \includegraphics[width=0.28\columnwidth]{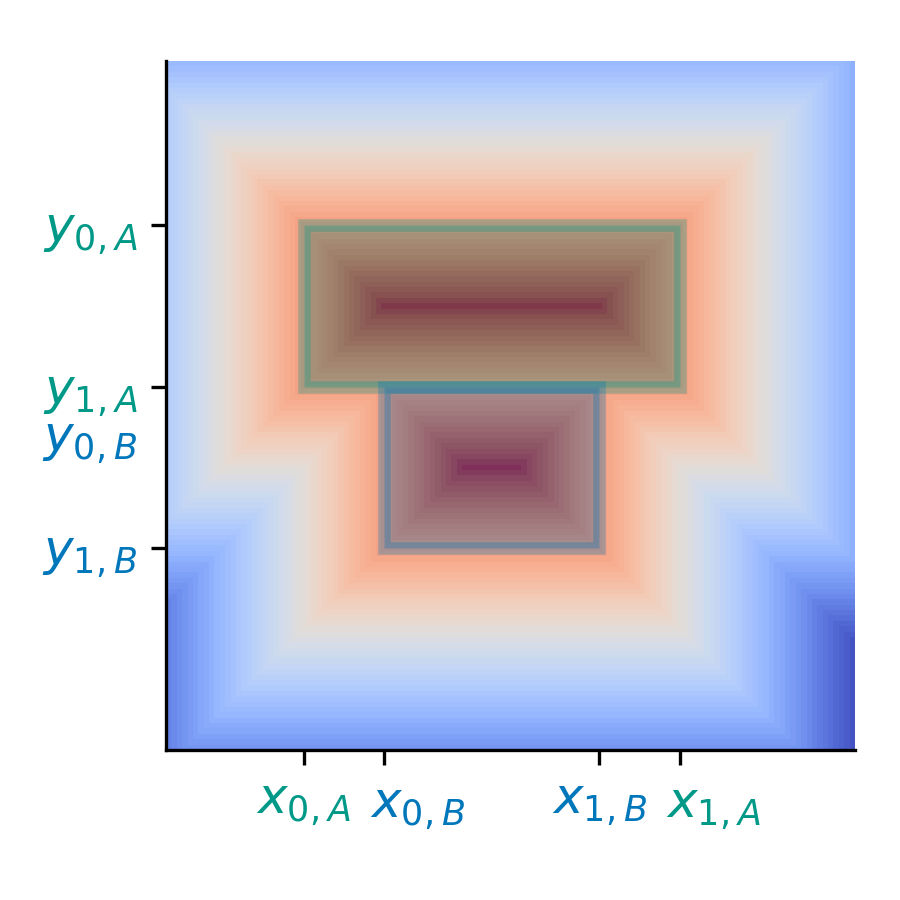}}
  \quad
  \subfloat[Door merge]{
  \includegraphics[width=0.28\columnwidth]{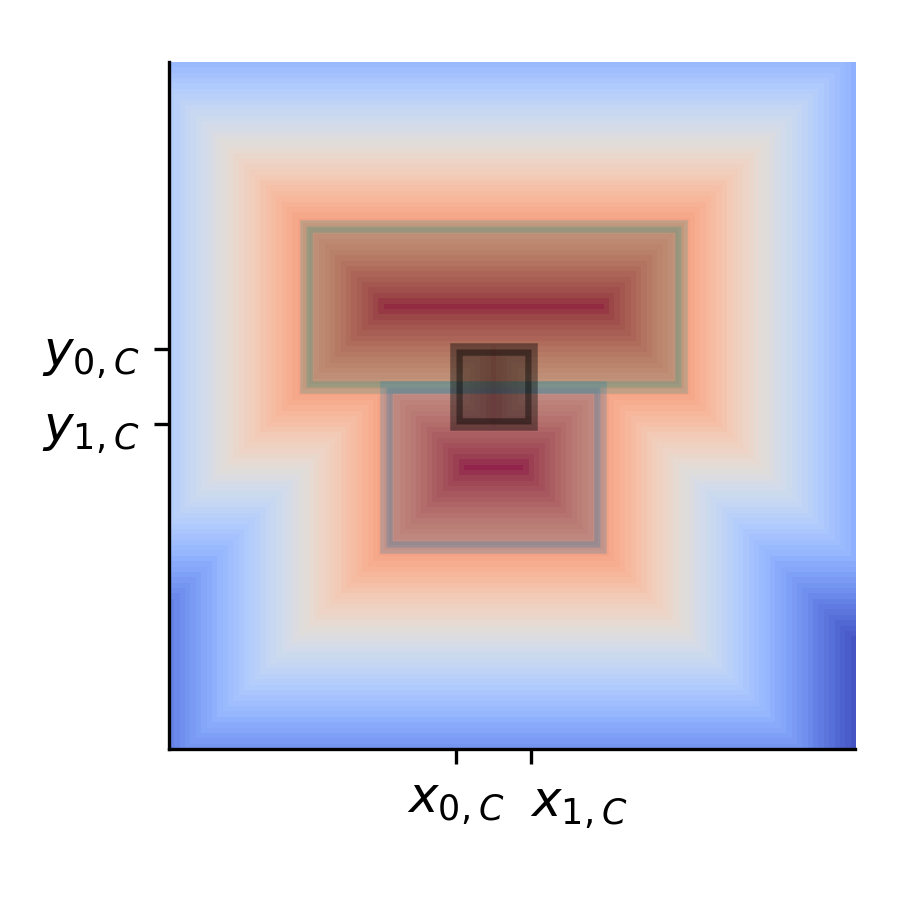}}
  \quad
  \subfloat[Non-rectangular]{
  \includegraphics[width=0.28\columnwidth]{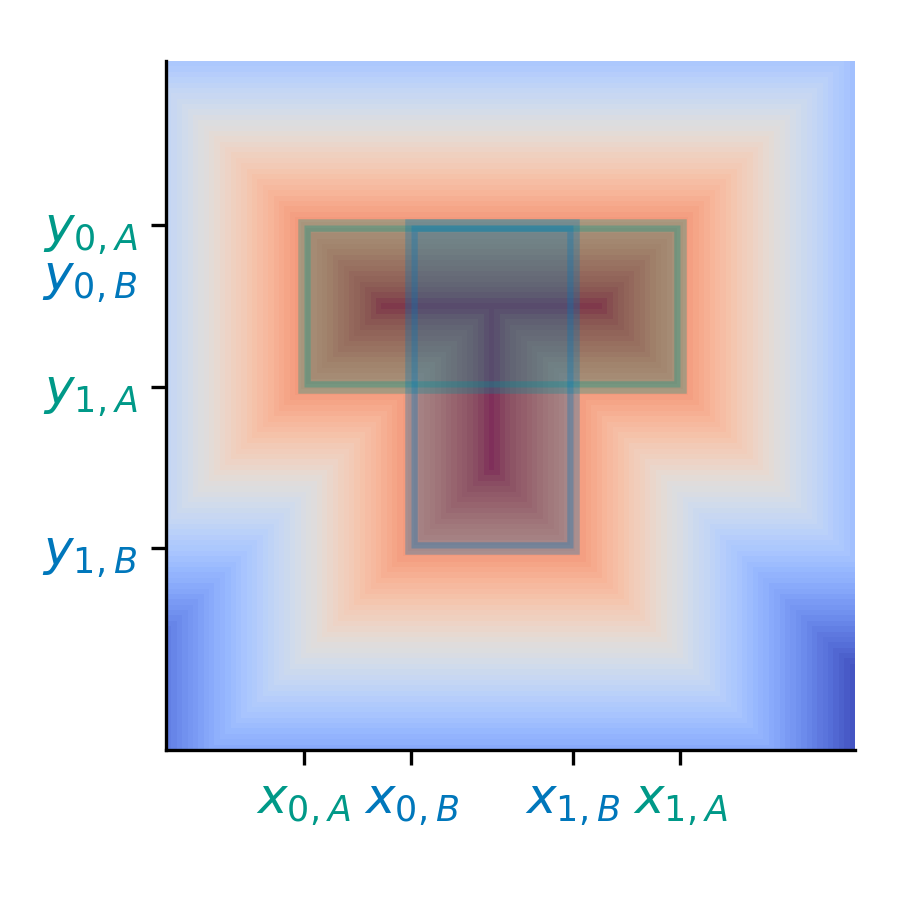}}
  \caption{Shape Merging Operation.}
  \label{fig:shape_merge}
\end{figure}
\myparagraph{Room (Multi-Box) Representation}
We represent a non-rectangular room as the merging of multiple overlapping boxes, and a doorway as the merging of a square-shaped box on the overlapping edges of two room boxes (Fig.~\ref{fig:shape_merge}). We can merge multiple SDFs using the $\max$ operation:
\begin{equation}\label{eq:final_sdf}
  f_{composite}(p) = \max_i \left(
              ~q^i \cdot (f_{2D}^i(p) + \gamma) - \gamma\right),
\end{equation}
where $i$ is the index of a box.

\subsection{Loss Functions}\label{sec:loss function}
We use two types of loss function to facilitate training: one based on TSDFs, the other on the box representation.

\myparagraph{Map-Based Loss}
We define a series of losses that compare TSDF representations. Thanks to our parametric TSDF representation (\Cref{subsec:sdf_from_box}), our approach  can use  raw occupancy maps for the training dataset (without the need for manual labeling of individual rooms), while still optimizing directly on the high-level box parameters.

\begin{figure}[tb]
  \centering
  \vspace{-2mm}
  \subfloat[]{
  \includegraphics[width=0.28\columnwidth]{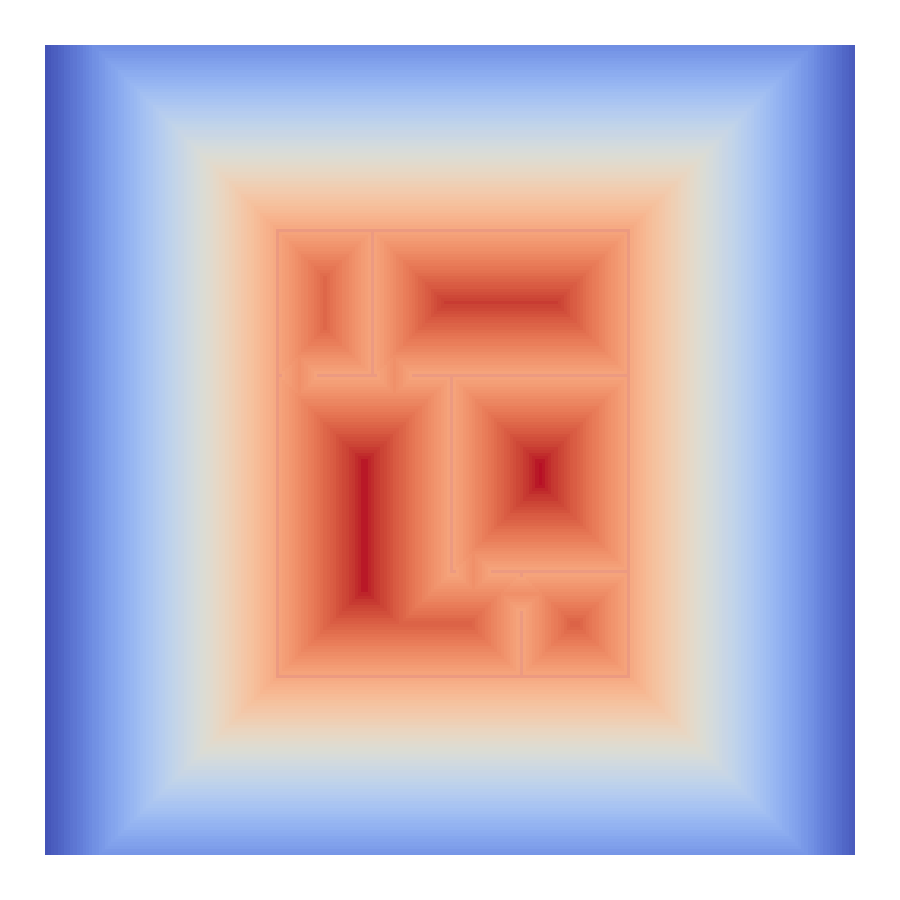}\label{fig:door_diamond_map}}
  \quad
  \subfloat[]{
  \includegraphics[width=0.28\columnwidth]{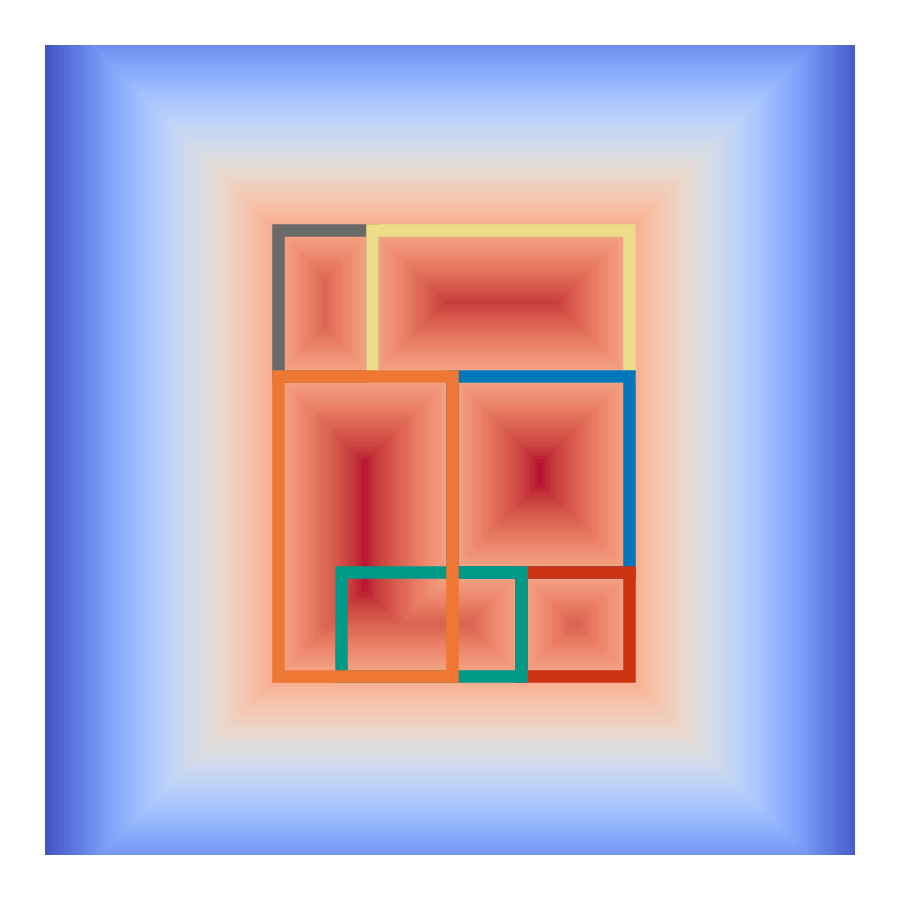}\label{fig:door_diamond_map_room}}
  \quad
  \subfloat[]{
  \includegraphics[width=0.28\columnwidth]{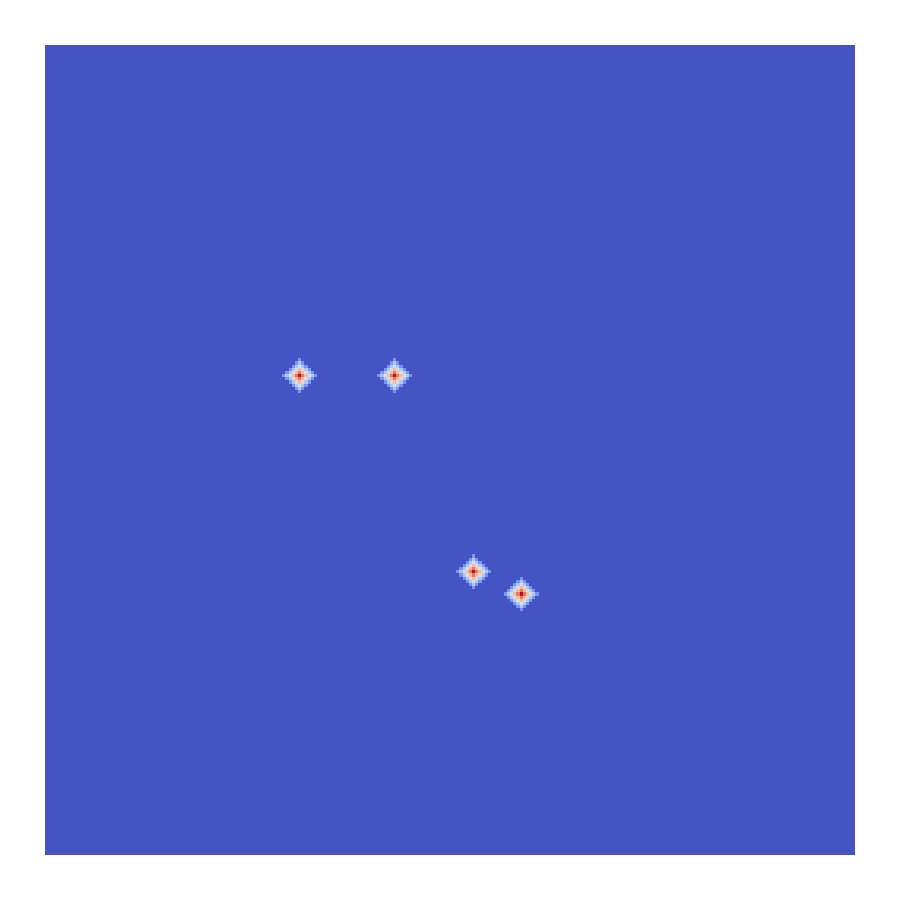}\label{fig:door_diamond_diff}}
  \caption{\protect\subref{fig:door_diamond_map} Ground truth TSDF with boundary overlaid in black, \protect\subref{fig:door_diamond_map_room} Predicted room TSDF, \protect\subref{fig:door_diamond_diff} Difference between \protect\subref{fig:door_diamond_map} and \protect\subref{fig:door_diamond_map_room} highlights the door locations.}
  \label{fig:door_diamond}
\end{figure}

We start with a basic $L_2$ loss over pixels $p$:
\begin{equation}
  l^p_{tsdf}(y_{tsdf},\hat{y}_{tsdf}) = \norm{y_{tsdf} -\hat{y}_{tsdf}}_2,
\end{equation}
where $y_{tsdf}$ is the ground truth TSDF, and $\hat{y}_{tsdf}$ is the predicted TSDF represented as~\eqref{eq:final_sdf}. We also introduce an auxiliary boundary loss
\begin{equation}
  l^p_{tsdf,W}(y,\hat{y}) = \norm{y_{tsdf} -\hat{y}_{tsdf}}^W_2,
\end{equation}
to emphasize accurate wall predictions, where the superscript $W$ indicates that pixels are masked by the ground truth wall region.
Empirically, the losses above do not capture doors well (due to their small size), in the sense that errors in the prediction of doors produce changes that are much smaller than errors in the prediction of rooms. We observe that the difference between the ground truth TSDF and the room-boxes-only TSDF, noted as $\tilde{y}_{tsdf}^{doors}$, highlights the locations of the doors (Fig.~\ref{fig:door_diamond}). We therefore define a hierachical loss that, assuming an accurate prediction of the rooms, focuses on doors:
\begin{equation}\label{eq:loss}
    \begin{aligned}
      l^p_{\Diamond doors}(p) &=  \norm{\tilde{y}_{tsdf}^{doors} - \max_i~\hat{y}_{\diamond}^{doors}(p,c^{i},s^{i})}^{\Tilde{W}}_2, \\
    \hat{y}_{\diamond}^{doors}(p,c^{i},s^{i}) &= ReLU(s^{i}/2 - ||p-c^{i}||_1),
    \end{aligned}
\end{equation}
where $\hat{y}_{\diamond}^{doors}$ is a TSDF with $c^{i}$ and $s^{i}$ being the centroid and dimension of a door box. 
The loss has increased penalties applied to door predictions on the region (denoted $\tilde{W}$) where the predicted TSDF $\hat{y}_{tsdf}$ is near zero (indicating walls) and the walls of the input.
The total map-based loss is
\begin{equation}
  \mathcal{L}_{TSDF} = \frac{1}{P} \left( \sum_{p} l^p_{tsdf}(y,\hat{y}) +l^p_{tsdf,W}(y,\hat{y}) + l^p_{\Diamond doors}(p)\right),
\end{equation}
where $P$ is the total number of pixels.

\myparagraph{Box-Based Loss} We include a self-supervised IOU loss to measure overlap between room boxes (denoted as $\cV$):
     \begin{equation}
        \mathcal{L}_{IoU}(\cV) =  \frac{1}{M (M-1)}\sum_{A\in \cV}\sum_{B\in \cV/A} IoU(A, B).
    \end{equation}

The gate loss on the set of existence probability of boxes $\mathcal{Q}$ is proposed as
\begin{equation}
        \mathcal{L}_{Gate}(\mathcal{Q}) = \frac{1}{\abs{\mathcal{Q}}}(\sum_\mathcal{Q} q^i (1 - q^i) + \sum_\mathcal{Q} q^i),
\end{equation}
where the first term is a bi-modal regularizer to force $q^i$ to be either $0$ or $1$, and the second term emphasizes sparsity.

\myparagraph{Final Loss}
The final loss is
\begin{equation}
  \begin{aligned}
  \mathcal{L}_{total}(\hat{y},y,\cV,\mathcal{Q}) =~& \mathcal{L}_{TSDF}(\hat{y},y)
  + \mathcal{L}_{IoU}(\cV) + \mathcal{L}_{Gate}(\mathcal{Q}).
  \end{aligned}
\end{equation}

\subsection{Topological Graph Generation}
A topological graph $\hat{\mathcal{G}}_{t}^{topo}$ is generated from the model outputs as follows:
\begin{enumerate*}
  \item The rooms $\cV$ in the environment are identified from the predicted room boxes, where a rectangular room is represented as a single node, and a non-rectangular room is represented as multiple overlapping box nodes with edges added between each pair without door information. Each node contains the location and dimension of the box. A pair of room boxes are connected with an edge if there exists a door connecting them.
  \item The doors are identified from the predicted door boxes. Each door contains the location, dimension and the connected room pair of the box. An edge is added to the graph if the points on opposite sides of the candidate door box are traversable within the door region, based on the accumulation of $\hat{M}_{t-1}^{topo}$ and $M_{t}^{laser}$.
\end{enumerate*}

\subsection{Dataset}
We used environments from the RPLAN dataset~\cite{RPLAN2019}, which contains 80k residential building floorplans. The TSDF for each environment was generated using the chamfer distance transform~\cite{borgefors1986distance}. We selected $400$ environments (each with five rooms) for training, and $41$ for testing. Note that the choice of five rooms was made to keep the setting manageable, allowing for easy interpretation of results. Preliminary results (not shown for space reasons) indicate our architecture can handle more complex environments with more rooms.

The robot was driven from random positions to implement standard frontier exploration in unknown environments to create the dataset. Each input and ground truth TSDF was obtained by cropping a $128\times 128$ local accumulated occupancy map. The model was initially trained on the occupancy maps, followed by fine-tuning with a dataset consisting predicted environments concatenated to the measurements.

\begin{figure*}[htbp!]
  \centering
  \includegraphics[width=1.82\columnwidth]{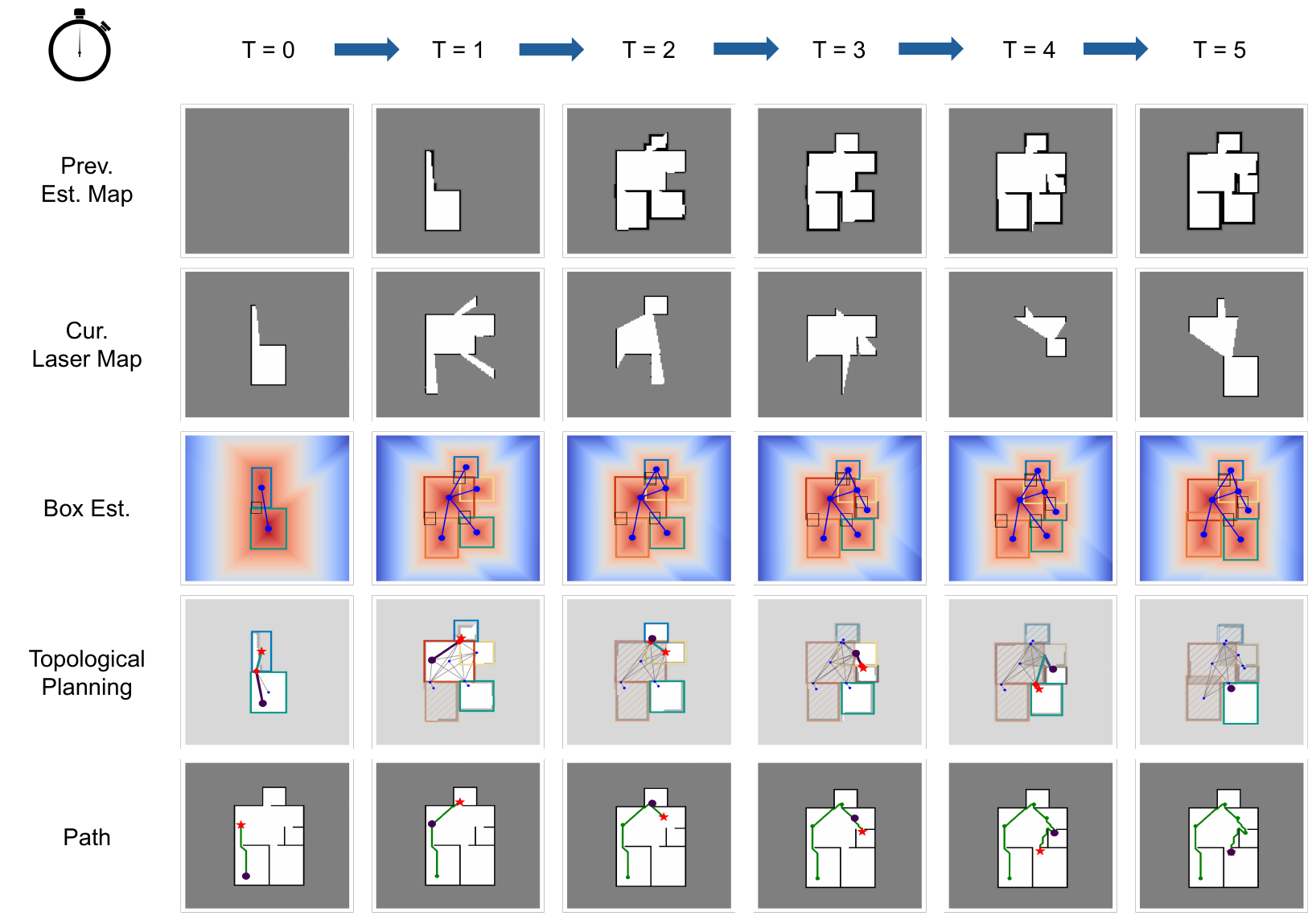}

   \caption{The progression of the agent exploration over time. It starts by instantiating the last estimated graph (row 1), then collects the laser measurement (row 2), and centers them to the network. The estimated room and door boxes (row 3) are used to construct a pose graph and plan the next move (row 4, shading indicates the room has been visited, overlaid with the predicted map). On row 4 and 5, the purple dot is the robot pose, red star is the point-to-go and red diamond is the nearest door. On the ground truth floorplan (row 5), green dots are the pose history of laser measurements.}
\label{fig:system_workflow}
\end{figure*}

\section{Graph-based Exploration}
In this section, we describe our algorithms for semantic exploration (Fig.~\ref{fig:system_workflow}) on graphs. 


\begin{figure}[htbp!]
  \centering
  \vspace{-2mm}
  \subfloat[Topological $\cG^{topo}$]{
  \includegraphics[width=0.28\columnwidth]{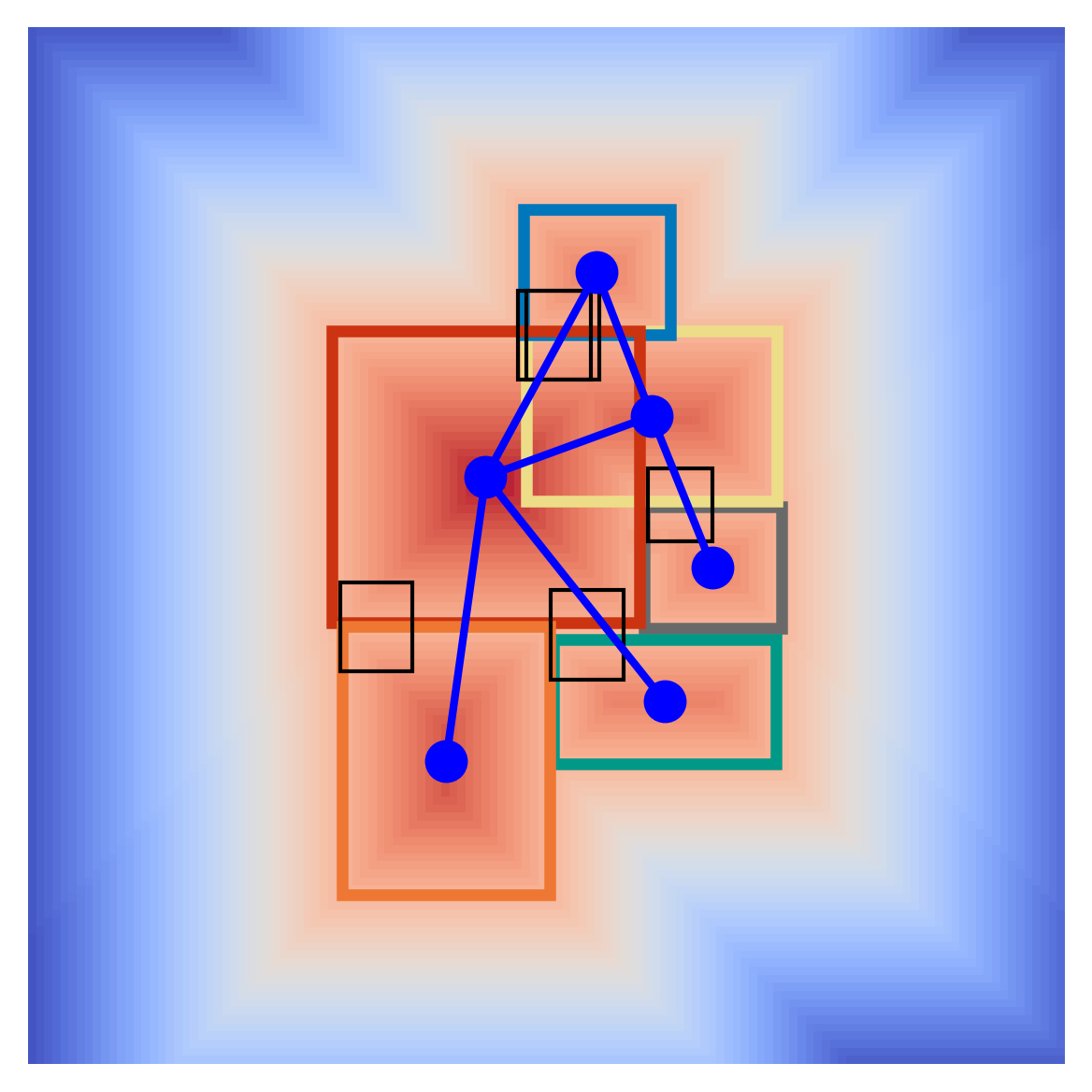}\label{fig:graph_generation_topological}}
  \quad
  \subfloat[Planning $\cG^{nav}$]{
  \includegraphics[width=0.28\columnwidth]{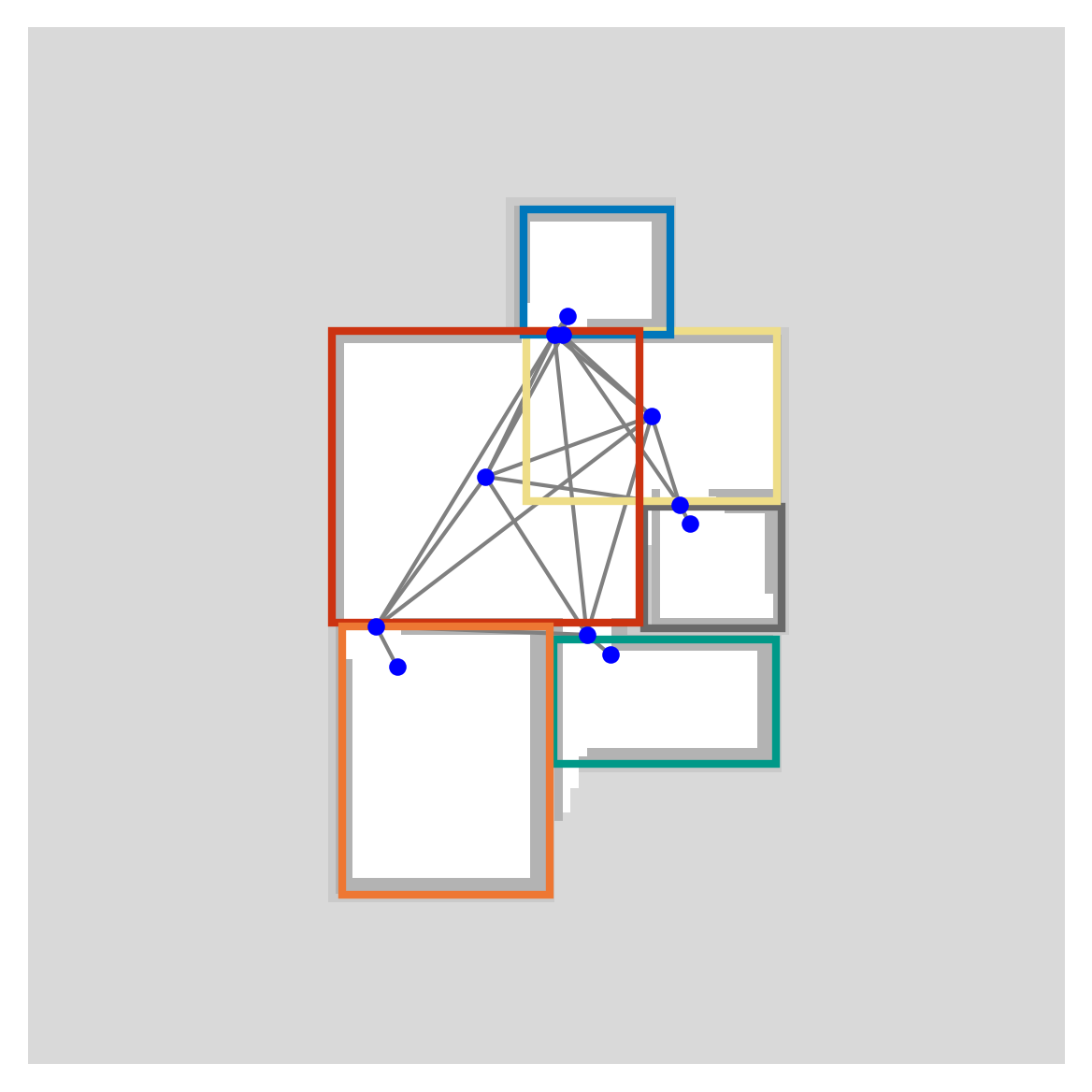}\label{fig:graph_generation_planning}}
  \quad
  \subfloat[Extended $\cG^{nav}$]{
  \includegraphics[width=0.28\columnwidth]{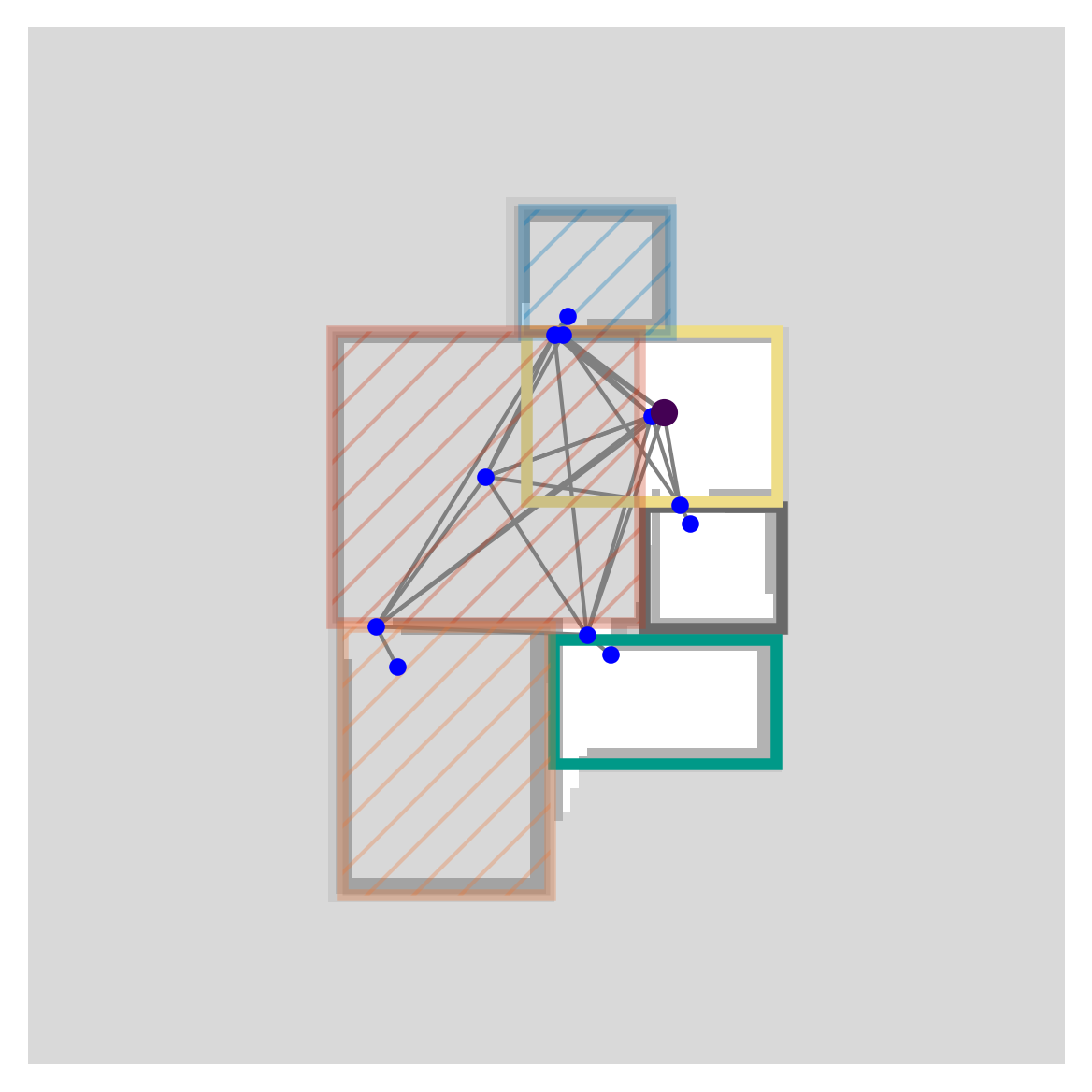}\label{fig:graph_generation_planning_pose}}
  \caption{Planning graph generation.}
  \label{fig:graph_generation}
\end{figure}

\subsection{Planning Graph Generation}
While the topological graph $\cG_t^{topo}$ (Fig.~\ref{fig:graph_generation_topological}) is a compact representation of the environment, we post-process it as $\hat{\mathcal{G}}_t^{nav}$ (Fig.~\ref{fig:graph_generation_planning}) to aid path planning. Specifically, we use the following steps:
\begin{enumerate*}
  \item Add a node for each door, connecting it with edges to the corresponding room pair.
  \item Separate doors in the same room are connected with an edge.
  \item Each rectangular room node has a position attribute at a point interpolated between the room centroid and its associated door centroid. We implement the interpolation by assuming that by entering a close-to-door region in a small room, the robot can capture the key features of the room (including itself and the interconnecting area) and make a reliable prediction.
\end{enumerate*} The planning graph is further extended to dynamically connect the robot node with the room-box node it resides in, as well as with the neighboring box nodes (Fig.~\ref{fig:graph_generation_planning_pose}).

\subsection{Semantic Exploration}\label{subsec:semantic_exploration}

We propose two different exploration strategies, one greedy and one that considers long-term performance using the global environment structure hallucinated from our model. We first identify previously visited nodes where measurements were taken. Both algorithms iteratively update $\hat{\mathcal{G}}_t^{nav}$  with new measurements once a selected frontier has been reached, and terminate when there are no more unvisited rooms.

\begin{lenumerate}[series=methods]{C}
  \item\textbf{Greedy strategy:}
  The robot uses the closest unvisited room on $\hat{\mathcal{G}}_t^{nav}$ as the interim goal $g_t$ .

  \item\textbf{Receding Horizon (RH) strategy:}
  At each iteration, the robot first plans the shortest path on $\hat{\mathcal{G}}_t^{nav}$ which visits all unvisited rooms at least once using Dynamic Programming. Then the robot selects the first room node on the generated path as the interim goal $g_t$.
\end{lenumerate}

\section{Simulations}
\subsection{Simulation Setup}
The simulations for the training dataset and  algorithm evaluations use PseudoSLAM~\cite{HouseExpo2019}. The laser scanner has a $360^{\circ}$ field of view and \unit[$9$]{m} range. All environments were represented as 2D occupancy maps with a resolution of \unit[$0.14$]{m} per pixel.

\subsection{Baselines}
We compare against two geometric-map based frontier exploration methods, where a frontier pixel is defined as the centroid of the segment that separates known regions from unknown regions. To give a fair comparison with our proposed method, the geometric map only accumulates laser measurement when a selected frontier has been reached.

\begin{lenumerate}[resume=methods]{C}
  \item\textbf{Traditional Frontier Exploration:}
  Candidate frontier pixels are selected using the method from~\cite{frontier_exploration2017}. Specifically, we define a reward $R(p_{f})$ that combines path cost and information gain for each frontier candidate pixel $p_{f}$:
  \begin{equation}\label{eq:frontier_revenue}
    R(p_{f}) = \lambda I(p_{f}) - \norm{p_{f}-p_t}_2,
  \end{equation}
  where $\|\cdot\|_2$ indicates the standard Euclidean metric,  $I(p_f)$ is the information of the frontier pixel defined as the number of unknown pixels within the range of the sensor around the pixel, and $\lambda$ is a tuning weight. The frontier candidate with the highest reward is selected as the next goal point. The process is repeated until there are no more frontier pixels.

  \item\textbf{Frontier Exploration with Map Completion:}
  We also consider a data-driven baseline~\cite{shrestha2019learned} where the candidate frontier pixels are selected based on the completed geometric map. We complete the partial map using the model trained in~\cite{wang2023more}. We find all the connected pixels that are unknown in the current map $M^{occ}_t$ but are free according to the completed map $\hat{M}^{occ}_t$; the number of such pixels is used as $I(p_{f})$.
\item\textbf{Hybrid Strategy:} For completeness, we add a strategy using the topological graph for exploration, while using the entire geometric map $M^{occ}_t$ for graph predictions.
\end{lenumerate}
\subsection{Evaluation of Semantic Exploration Performance}
We ran the algorithms on $41$ test environments, from three random initial positions each. To evaluate performance we define the following metrics averaged over all runs.
\begin{lenumerate}{R}
        \item Total steps: number of pixel traversed to complete.  \label{metrics_steps}
        \item Number of measurement updates. \label{metrics_iter}
\end{lenumerate}
We consider three additional metrics that are not directly related to the exploration but are useful to assess the differences between our algorithms and the baselines.
\begin{lenumerate}{M}
        \item Map memory \label{metrics_map}
        \item Structural Similarity Index Measurement (SSIM) between the final TSDF and the ground truth map. \label{metrics_ssim}
        \item Hamming loss: pixel-wise classification accuracy of the final map. \label{metrics_hamming}
\end{lenumerate}

\setlength{\tabcolsep}{4.5pt}
\begin{table}[tb]
  \vspace{1.5mm}
  \caption{Evaluation results. $\downarrow$ indicates that smaller values imply better performance. $N=256$ and $M=6$ here.}
  \label{table:method_comparison}
  \centering
  \begin{tabular}{llllll} 
          \toprule
          &\multicolumn{1}{c}{\ref{metrics_steps} $\downarrow$}
          &\multicolumn{1}{c}{\ref{metrics_iter} $\downarrow$}
          &\multicolumn{1}{c}{\ref{metrics_map} $\downarrow$}
          &\multicolumn{1}{c}{\ref{metrics_ssim} $\uparrow$}
          &\multicolumn{1}{c}{\ref{metrics_hamming} $\downarrow$}\\
          \midrule
          BoxMap Greedy (C1) & 114.9 & 5 & $\mathcal{O}(M^2)$ & 0.96 &  2.5\%\\
          BoxMap RH (C2) & 115.2 & 5 & $\mathcal{O}(M^2)$ & 0.96 &  2.4\%\\
          Occ. Standard (C3)  & 166.4 & 4.6 & $\mathcal{O}(N^2)$ & 1 &  0\\
          Occ. NN (C4)  & 143.6 & 4.8 & $\mathcal{O}(N^2)$ & 1 &  0\\
          Occ. RH (C5) & 111 & 4.9 & $\mathcal{O}(N^2)$ & 0.98 &  1.3\%\\
          \bottomrule
  \end{tabular}
\end{table}

\begin{figure}[tb]
  \centering
  \vspace{-5mm}

  \subfloat[Total steps]{
  \includegraphics[width=0.42\columnwidth]{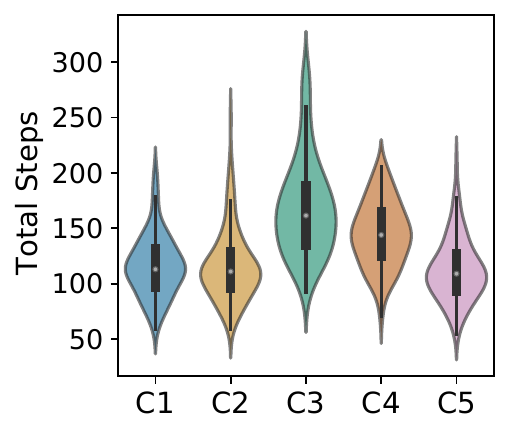}}
  \qquad
  \subfloat[Measurement updates]{
  \includegraphics[width=0.42\columnwidth]{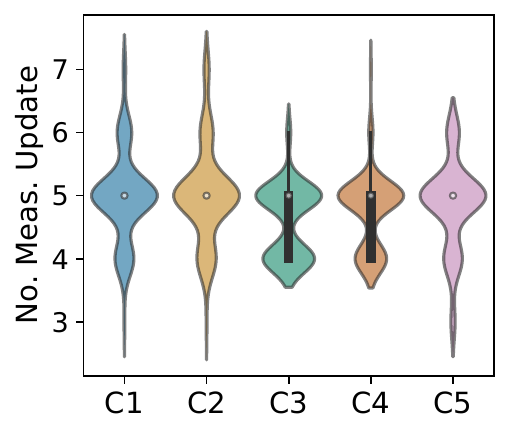}}
\caption{Violin plot over all runs of (a) total steps and (b) total measurement updates. Violin plots illustrate data distributions by superimposing kernel density plots onto box plots.}
\label{fig:method comparison}
\end{figure}
Our results, summarized in Table~\ref{table:method_comparison} and Fig.~\ref{fig:method comparison}, show that our method is superior to the baseline, yielding a more compact map representation as well as shorter trajectories. This indicates that our method can leverage semantic prior knowledge learned from data to create a higher-level representation from low-level partial measurements, which result in improved performance; such information can not be directly obtained from any geometric-based frontier methods. Moreover, despite the reduction in storage, the maps that can be reconstructed by our methods show only a minor reduction in accuracy with respect to geometric maps (as indicated by metrics \ref{metrics_ssim} and \ref{metrics_hamming}). It should be noted that our method discards low-level geometric information other than the current measurement. This is achieved by converting $\hat{\cG}^{topo}_{t-1}$ to $\hat{\cM}^{topo}_{t-1}$, and using $\hat{\cM}^{topo}_{t-1}$ as the new model input. While there are potential prediction and conversion errors, the model may miss some visited semantic entities in the long horizon. We have observed, however, that the model is able to correct the error if the affected area is revisited (at the cost of longer trajectories).

\subsection{Simulation in Gazebo}
\begin{figure}[tb]
  \centering
  \vspace{-2mm}
  \subfloat[Init. Pred.]{
      \includegraphics[width=0.3\columnwidth]{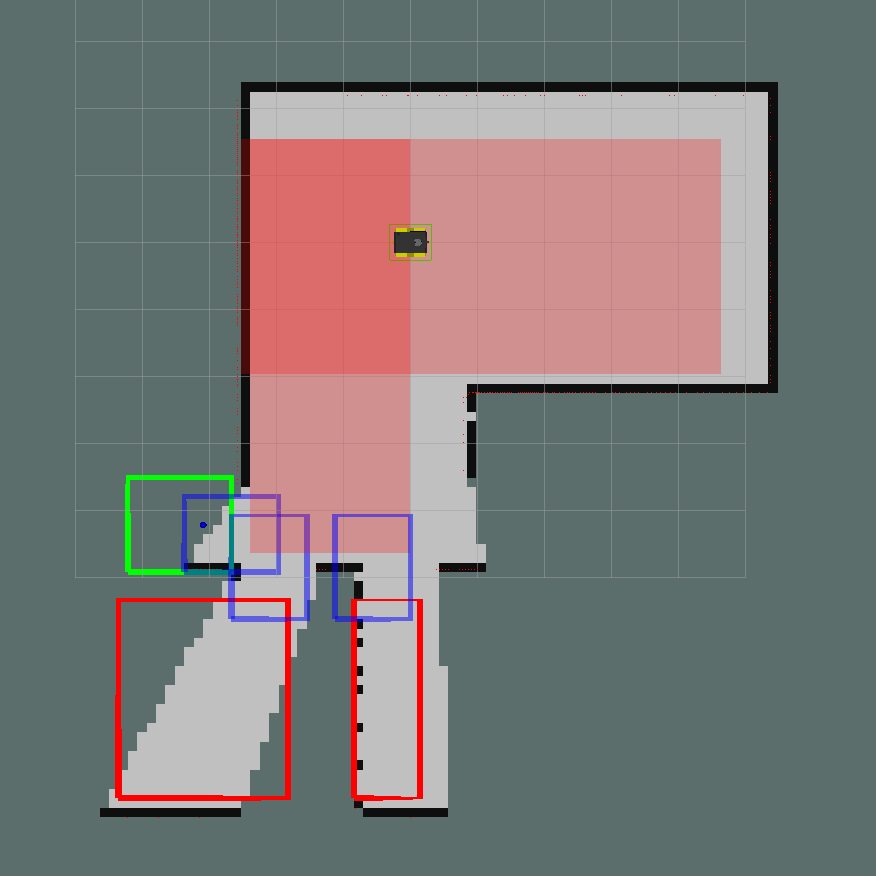}
      \label{fig:gazebo_sim_first}
    }
    \quad\quad\quad
    \subfloat[Final Pred.]{
      \includegraphics[width=0.3\columnwidth]{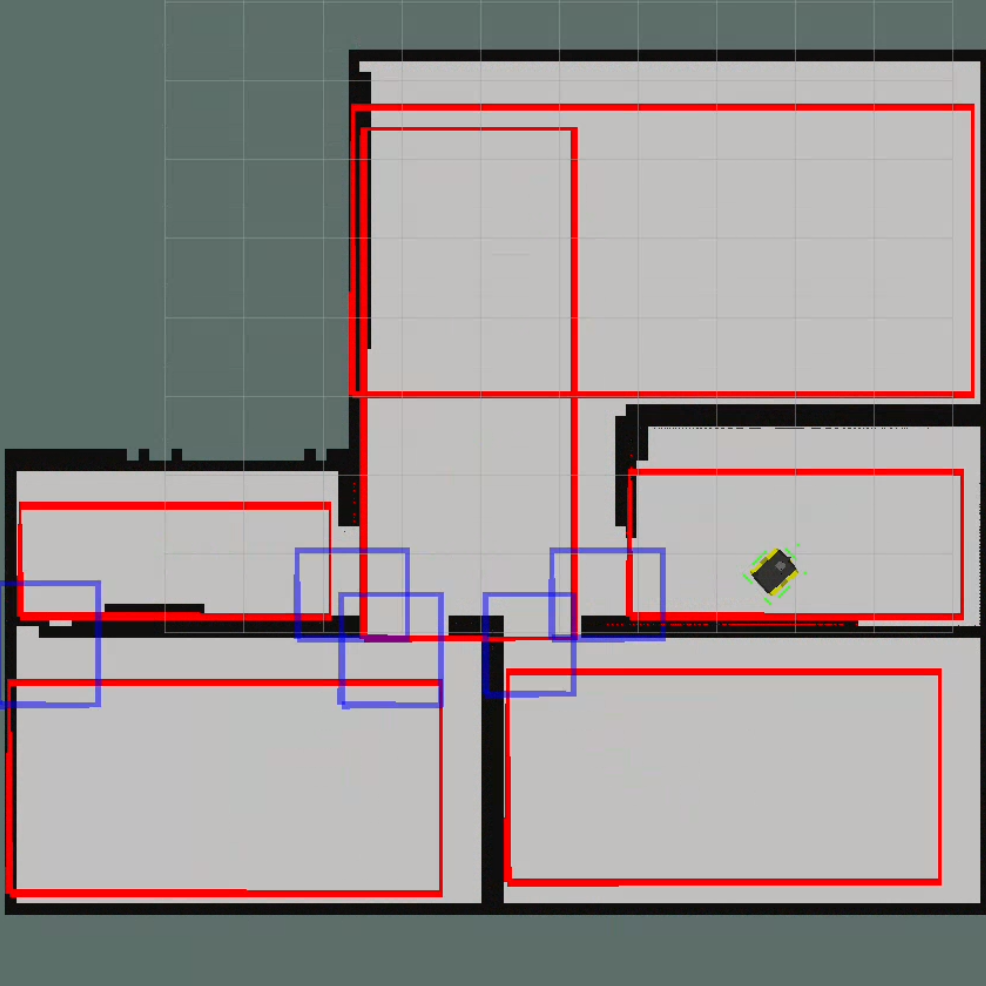}
      \label{fig:gazebo_sim_final}}
  \caption{ROS Simulation: \protect\subref{fig:gazebo_sim_first} The initial LiDAR scan of the Jackal with room (red and green rectangles, green indicated the selected frontier) and door (blue rectangle) predictions, shaded-in red indicates the room that robot starts in. \protect\subref{fig:gazebo_sim_final} Final predictions after having explored the entire map.}
  \label{fig:gazebo_sim}
\end{figure}

We further test our exploration algorithm through Robot Operating System (ROS) using the Gazebo simulator with a Jackal robot (Clearpath Robotics) equipped with a LiDAR sensor, leveraging the \texttt{gmapping} package to accumulate the laser measurements.
Accounting for the robot's footprint, to avoid collisions with obstacles, we add waypoints near doors to traverse smoothly these narrow regions. Fig.~\ref{fig:gazebo_sim_first} demonstrates an initial random placement of the robot in an apartment layout, the initially constructed graph and the selected frontier. The algorithm is repeated until all the rooms have been visited. Fig.~\ref{fig:gazebo_sim_final} shows the graph $\hat{\cG}^{topo}$ when the algorithm terminates. The result shows that BoxMap is robust to imperfect box predictions due to noisy sensor and odometry measurements. Simulations in multiple environment layouts (not shown due to space constraints) confirmed these findings.
\section{CONCLUSIONS}
We proposed a CNN-transformer-based architecture to learn and update topological information of the environment from low-level measurements, which significantly reduces the map storage in navigational execution. To facilitate the training process, we proposed to learn from a TSDF-map instead of box vertices and demonstrated its strength on learning semantic entities and relations. Through simulations we demonstrated that our graph-based semantic exploration algorithm achieved better performance compared to geometric-map-based frontier exploration algorithms.
While our algorithms assumed both perfect odometry and sensing, our simulations in Gazebo indicated that it can be adapted to realistic settings.

Future work includes testing on real hardware and improving the robustness of the architecture by either training with clustered environments or combining with obstacle removal algorithms, and exploring the use of DETR extensions (e.g. Deformable DETR) that are potentially more robust to small box detection.  A direct output feedback controller in polygonal environment will also be considered.


\bibliographystyle{biblio/IEEEtranNoDash}

\bibliography{biblio/IEEEfull,biblio/IEEEConfFull,biblio/OtherFull,
biblio/semantic_nav,biblio/object_detection,biblio/tron,biblio/websites}

\end{document}